\documentclass[10pt,twocolumn,letterpaper]{article}
\usepackage{wacv} % Camera-ready is the default in the supplied WACV 2027 style
\usepackage{graphicx,amsmath,amssymb,booktabs,tabularx,array}
\usepackage{placeins}
\definecolor{wacvblue}{rgb}{0.21,0.49,0.74}
\usepackage[breaklinks,colorlinks,allcolors=wacvblue]{hyperref}
\hypersetup{pdfauthor={Syeda Hareem Madani, Sumra Khan, Noureen Bibi, Adam Rafiq Jeraj, Anas Zafar, Inad Alqurashi, Rizwan Qureshi},pdftitle={Beyond Anatomy: Interpreting Graph-Based ASD Classification Across ABIDE Settings}}
\def\confName{WACV}
\def\confYear{2027}
\title{Beyond Anatomy: Interpreting Graph-Based ASD Classification\\Across ABIDE Settings}
\author{%
Syeda Hareem Madani\textsuperscript{1}, Sumra Khan\textsuperscript{2,$\ast$}, Noureen Bibi\textsuperscript{1},\\
Adam Rafiq Jeraj\textsuperscript{3}, Anas Zafar\textsuperscript{4}, Inad Alqurashi\textsuperscript{5},\\
Rizwan Qureshi\textsuperscript{6}\\[4pt]
{\small \textsuperscript{1}Salim Habib University \quad \textsuperscript{2}GU Tech, Al Ghazali University}\\
{\small \textsuperscript{3}National University of Singapore \quad \textsuperscript{4}The University of Texas MD Anderson Cancer Center}\\
{\small \textsuperscript{5}Taif University \quad \textsuperscript{6}Massachusetts General Hospital}\\[2pt]
{\small $\ast$ Corresponding author: Sumra Khan (\href{mailto:sumra.khan@agu.edu.pk}{sumra.khan@agu.edu.pk})}
}
\begin{document}
\maketitle
\begin{abstract}
Graph neural networks can classify resting-state functional MRI connectomes, but a high score on one configuration leaves open which representation choices matter and whether the result supports broader clinical use. We examine this problem through two ABIDE-I analyses. On a balanced 400-subject subset, an AAL-based graph convolutional network (73.3\%), an MSDL-based configuration (84.0\%), and a five-model graph attention ensemble (95.0\%) report high nominal performance. However, we show that this 95.0\% result, which substantially exceeds typical multi-site ABIDE benchmarks (65–70\%), stems from a single site-stratified test split of 60 subjects and does not reflect out-of-distribution generalization to unseen acquisition centers. In a separate 1,035-subject analysis with different preprocessing and parcellation, BrainGAT reaches 58.26\% mean accuracy, below an SVM at 69.37\%. We interpret these as comparisons of complete pipelines and retain both the favorable ensemble result and the negative broader-cohort result. Analysis of the existing summaries shows that atlas substitution changes the number of possible region pairs approximately ninefold, that accuracy on 60 test subjects changes by 1.67 percentage points per prediction, and that the best accuracy and AUC belong to different broader-cohort classifiers. Regional attribution on the baseline GCN (not the 95\% GAT ensemble) highlights default-mode-network features but reflects classifier reliance rather than validated clinical biomarkers. Unresolved metric aggregation and diagnosis-aware harmonization limit the findings to exploratory classification evidence.
\end{abstract}
\vspace{-10mm}
\section{Introduction}
Resting-state functional MRI (rs-fMRI) represents brain activity through regional time series and their statistical relationships. For autism spectrum disorder (ASD), distributed and individually variable connectivity motivates classifiers that integrate information across regions~\cite{hahamy2015idiosyncratic}. ABIDE provides a multi-site resource for studying such models~\cite{di2014autism}. The translational challenge is to determine which evidence supports prediction for new people and acquisition conditions, rather than treating a favorable benchmark score as a clinical endpoint~\cite{plitt2015functional}.

Graph representations make this challenge concrete. A parcellation determines the nodes and signals available to the model. Correlation estimation and sparsification determine candidate relationships. Graph convolutional networks (GCNs) aggregate neighborhood features, and graph attention networks (GATs) learn adaptive neighborhood weights~\cite{kipf2017semi,velickovic2018graph}. Changing any one of these steps can alter the input distribution, statistical complexity, or inductive bias. Changes made together are therefore properties of a complete pipeline, even when an atlas gives the configuration its name.

The performance escalates across an AAL GCN, an MSDL GCN, and a GAT ensemble in an experiment involving 400 subjects in a balanced cohort. In contrast, a separate 1,035-subject analysis prefers conventional vectorized classifiers over a graph attention/convolution model. Both observations are valid: the first identifies an encouraging configuration in a selected cohort; however, the second provides no evidence of a general graph-model advantage. Their contrast is not a transfer experiment on the same trained model because the pipelines differ.

The paper examines three questions. \textbf{Q1: What does the configuration ranking demonstrate about representation?} The simultaneous architectural changes are documented in this paper, and we quantify the shift in available region pairs. \textbf{Q2: How much evidence do the reported scores provide?} We study within-setting rankings, the limits of the broader-cohort protocol, and test-set granularity. \textbf{Q3: What does regional attribution establish?} We interpret the provided saliency summary as a hypothesis about model reliance, distinct from predictive validity and clinical relevance.

Our contribution is an empirical case study with an analysis of existing results, rather than a new graph operator. The positive and negative findings are brought together into one account, which links each conclusion to the comparison that supports it and adds explicit calculations of evaluation sensitivity and representation size. All reported experimental scores are included. Calculations derived from those summaries are identified separately from measured results. This distinction is particularly relevant to clinical neuroscience applications in which subject, site, and preprocessing differences can change the meaning of a model's apparent success.

\section{Related Work}
\paragraph{Connectome-based classification.}
ABIDE supports the investigation of ASD-related variability across acquisition sites~\cite{di2014autism}. Deep learning studies have explored connectivity features and volumetric or connectome representations~\cite{heinsfeld2018identification,khosla2019ensemble}. Evaluation design highlights the importance of reproducible multi-site biomarkers~\cite{abraham2017deriving}, while the evidentiary requirements for a clinical biomarker extend beyond predictive connectivity patterns~\cite{plitt2015functional}. Accordingly, we contextualize the conclusions of studies that share a test set rather than reporting their performance as independent evaluations.

\paragraph{Parcellation and graph models.}
AAL defines anatomical regions~\cite{tzourio2002automated}, whereas MSDL uses dictionary learning to obtain spatial maps of spontaneous activity~\cite{varoquaux2011multi}. A multi-atlas approach has been investigated to explore complementary representations for ASD classification~\cite{liu2024made}. The application of graph convolutional networks (GCNs)~\cite{kipf2017semi} to ASD classification~\cite{anirudh2019bootstrapping} motivates graph-based modeling. Graph attention networks (GATs) introduce adaptive neighborhood weights~\cite{velickovic2018graph}. Our comparison does not establish that functional atlases universally outperform anatomical atlases, nor does it support conclusions across all possible configurations.

\paragraph{Interpretability.}
GNNExplainer identifies features relevant to a trained model and influential graph structures~\cite{ying2019gnnexplainer}. Such explanations describe model behavior but do not establish that the attributed features are causal, novel to ASD, or free from acquisition-related confounding. Previous observations involving default-mode connectivity~\cite{kennedy2008functional} provide context for interpreting our regional attributions rather than constituting independent validation of the classifier.

\section{Data and Evaluation Settings}
\label{sec:settings}

Table~\ref{tab:settings} distinguishes the two evaluation settings. Both analyses use ABIDE-I; therefore, the broader cohort does not constitute an independent external validation dataset. We do not treat their results as independent replications, and the overlap in subject identities between the analyses is not specified.

\begin{table*}[t]
\centering
\small
\caption{Comparison of the curated subset and broader ABIDE-I cohort across dataset composition, preprocessing, parcellation, evaluation, and modeling configurations. The two settings differ across multiple methodological factors and therefore do not isolate the effect of any single factor.}
\label{tab:settings}
\begin{tabularx}{\textwidth}{lXX}
\toprule
 & Curated Subset & Broader Cohort \\
\midrule
Subjects & 400: 200 ASD, 200 controls & 1,035: 505 ASD, 530 controls \\
Reported site groups & 17 & 20 \\
Preprocessing & Manual FSL workflow & CPAC, filt-noglobal strategy \\
Parcellation & AAL (116); MSDL (39) & Harvard--Oxford (111) \\
Evaluation & Site-stratified 280/60/60 split & Stratified five-fold cross-validation \\
Models & Two GCN configurations; GAT ensemble & SVM, logistic regression, MLP, BrainGAT \\
Graph sparsity & Top 20\% of connections in AAL workflow & Top 15\% of absolute correlations \\
Harmonization & No ComBat procedure specified & ComBat; site, age, sex, and diagnosis recorded \\
\bottomrule
\end{tabularx}
\end{table*}

\subsection{Curated Cohort and Partitioning}

No site-follow-up data were available for comparison with the experiments, and a common mapping of acquisition-site indicators was unavailable. The curated analysis includes 400 subjects, equally partitioned between the ASD and typically developing control groups. The reported age range is 7--64 years, and the ASD group is predominantly male.

The aggregate report does not provide exact sex counts, per-site subject counts, or age distributions by diagnosis. Consequently, demographic matching cannot be assessed from the available information.

The stated 70/15/15 subject split corresponds to 280 training, 60 validation, and 60 test subjects. Site stratification does not imply site exclusion: subjects from represented sites can occur in all three partitions. This distinction matters because evaluating on different subjects from represented sites is not equivalent to evaluating on an entirely unseen acquisition site.

The reported 95.0\% accuracy should therefore not be interpreted as evidence of domain generalization or clinical readiness. The 60 test subjects originate from represented sites rather than exclusively held-out acquisition sites. In literature-wide ABIDE-I benchmarks, performance under strict cross-site or cross-dataset validation has often been reported in the range of 60\%--70\%. Residual associations between acquisition characteristics and diagnosis cannot be ruled out by site stratification alone.

\subsection{Image Preprocessing}

The curated workflow is based on FSL~\cite{jenkinson2012fsl}. Structural processing begins with skull stripping of the MPRAGE image using BET, with a fractional intensity threshold of 0.5. The image is then affine-registered to MNI152 2-mm space using FLIRT with 12 degrees of freedom, followed by nonlinear registration to MNI152 2-mm space using FNIRT.

Functional preprocessing includes interleaved slice-timing correction, MCFLIRT motion correction, and boundary-based registration to the structural image. The processing description reports volume exclusion when mean displacement exceeds 3 mm and visual inspection of image alignment.

\begin{figure*}[t]
\centering
\includegraphics[width=0.85\textwidth]{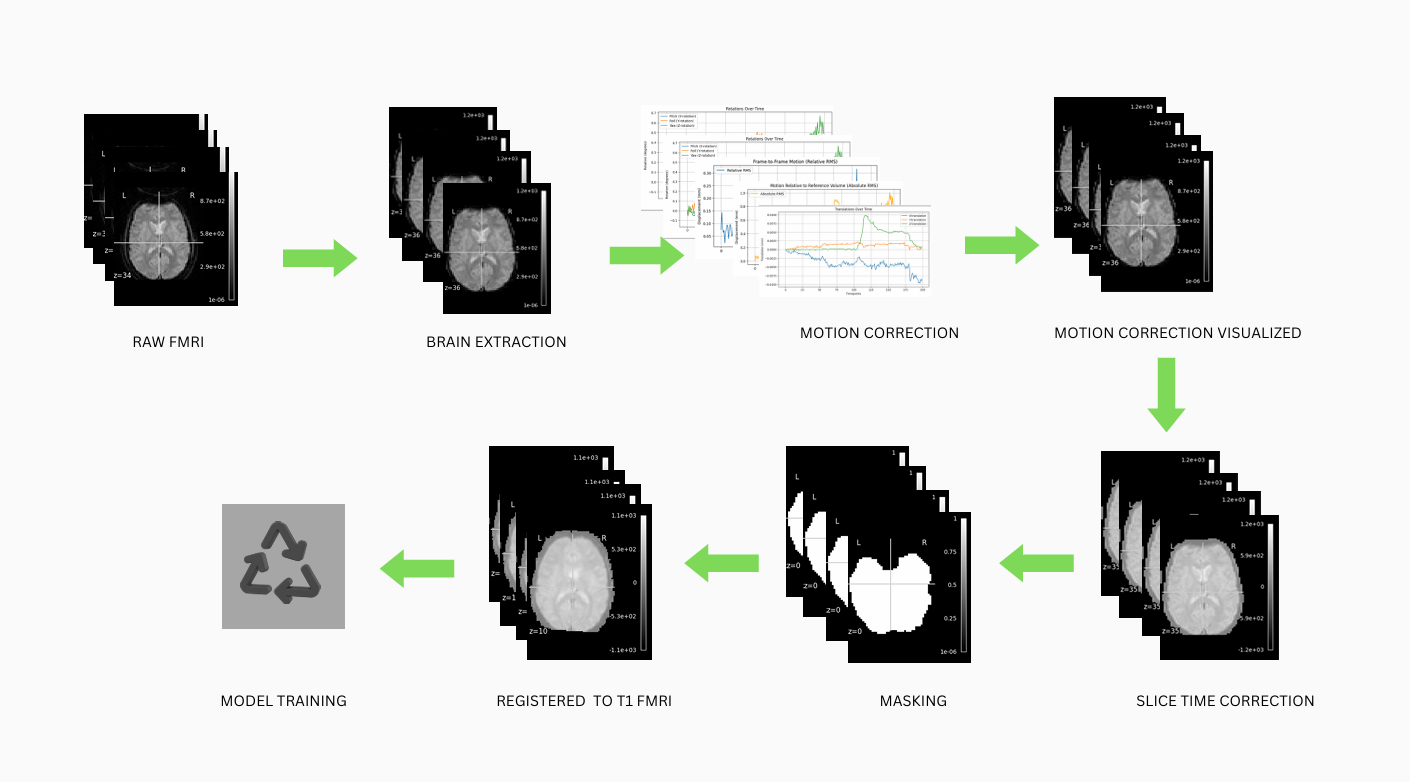}
\caption{Preprocessing workflow for the curated cohort before regional signal extraction and graph classification: (1) raw rs-fMRI acquisition, (2) BET brain extraction (threshold 0.5), (3) MCFLIRT motion correction and alignment inspection, (4) interleaved slice-timing correction, (5) binary masking, (6) boundary-based registration to structural T1 and MNI152 space, and (7) regional time-series extraction and graph model training.}
\label{fig:preprocessing}
\end{figure*}

Figure~\ref{fig:preprocessing} illustrates the image-preparation steps preceding regional signal extraction. However, quantitative quality-control summaries are not available. Reproducibility is consequently limited by incomplete reporting of motion metrics, nuisance regression, site-specific slice-timing procedures, and filtering settings.

\subsection{Broader-Cohort Analysis}

The broader analysis includes 1,035 subjects (505 ASD and 530 controls; 48.8\% ASD) across 20 acquisition-site groups. It uses data from the ABIDE Preprocessed Connectomes Project, with CPAC preprocessing (filt-noglobal strategy) and a Harvard--Oxford parcellation containing 111 regions. These data and features should not be interpreted as representing the same cohort configuration as the curated analysis.

The experiment uses Pearson correlation and tangent-space features, with ComBat harmonization. Site, age, sex, and diagnosis are recorded as batch-related or covariate variables in the reported harmonization procedure. Evaluation uses stratified five-fold cross-validation with a fixed random seed. However, no held-out-site split or independent deployment evaluation is reported.

\paragraph{Evaluation qualification.}
In the broader cohort (1,035 subjects), ComBat harmonization incorporated diagnosis alongside site, age, and sex. For strict label-blind evaluation, batch transformations, tangent-space reference matrices, and feature scalers must be estimated solely on training folds. Because the analyzed records do not confirm that transformations and model selection were strictly isolated to training folds, and because test-set feedback may have guided ensemble checkpoint selection, all broader-cohort metrics are cataloged as exploratory estimates with unresolved data leakage risks.

\section{Graph Representation and Models}
\subsection{Connectivity graphs}
Regional time series are extracted using AAL (116 regions) or MSDL (39 probabilistic regions). For regional signals $s_i$ and $s_j$, connectivity is represented by their Pearson correlation,
\begin{equation}
 C_{ij}=\frac{\sum_t(s_i(t)-\bar{s}_i)(s_j(t)-\bar{s}_j)}{\sqrt{\sum_t(s_i(t)-\bar{s}_i)^2}\sqrt{\sum_t(s_j(t)-\bar{s}_j)^2}}.
\end{equation}
Regions are graph nodes; correlations define candidate edge weights. AAL graph construction retains the top 20\% of the strongest connections. The MSDL workflow also uses correlation-based connectivity, but the exact treatment of signed edges, sparsification, and probabilistic-map extraction is not fully specified. We do not infer those choices from the broader-cohort configuration, which explicitly uses absolute correlations.

The node-feature tensor and self-loop handling are incompletely specified in the available implementation description. We distinguish this missing information from the known connectivity construction. The dimensional consequences of changing the atlas are analyzed in Section~\ref{sec:representation}.

\subsection{Curated-cohort architectures}
The AAL baseline has two graph convolutional layers with hidden widths of 64 and 32, ReLU activations, a dropout rate of 0.3, and global mean pooling. The MSDL GCN has three hidden layers of width 64 and batch normalization. These differences are retained explicitly in Table~\ref{tab:models}; they preclude an atlas-only interpretation of the accuracy difference.

The ensemble contains five independently trained GAT models. Each uses two attention layers with two heads, a hidden width of 64, a dropout rate of 0.5, global mean pooling, and 20\% edge dropout during training. Soft voting combines their predicted probabilities. For a subject graph $G$, the ensemble probability is
\begin{equation}
 p_{\mathrm{ens}}(y\mid G)=\frac{1}{5}\sum_{m=1}^{5}p_m(y\mid G).
\end{equation}
The ensemble is described in the functional-parcellation pipeline, but its exact atlas is not explicitly recorded in the architecture specification. We leave that field unspecified in Table~\ref{tab:models} rather than infer it from the GCN configuration.

\begin{table}[t]
\centering\small
\caption{Curated-cohort model configurations. GNNExplainer is a post-hoc explanation method applied to the baseline GCN, not the ensemble classifier.}
\label{tab:models}
\footnotesize\setlength{\tabcolsep}{3pt}
\resizebox{\columnwidth}{!}{%
\begin{tabular}{lccc}
\toprule
 & AAL GCN & MSDL GCN & GAT ensemble \\
\midrule
Regions & 116 (AAL) & 39 (MSDL) & 39 (MSDL) \\
Node features & 116-d profile & 39-d profile & 39-d profile \\
Self-loops & Included & Included & Included \\
Graph sparsity & Top 20\% abs. corr. & Top 20\% abs. corr. & Top 20\% abs. corr. \\
Hidden widths & 64, 32 & 64, 64, 64 & 64, 64 \\
Attention heads & -- & -- & 2 \\
Batch norm. & No & Yes & No \\
Dropout & 0.3 & NR & 0.5 (20\% edge drop) \\
Pooling & Mean & Mean & Mean \\
Augmentation & None & None & Gaussian ($\sigma=0.05$) \\
Models & 1 & 1 & 5 \\
\bottomrule
\end{tabular}}
\end{table}

\subsection{Training and augmentation}
The curated models are reported to use Adam with a learning rate of $10^{-3}$ for 100 epochs. Gaussian noise augmentation~\cite{shorten2019survey} is applied to correlation matrices for ensemble training, with a mean of zero and a standard deviation of 0.05. We use the stated training-only protocol: each of the 280 training subjects contributes the original graph and five noisy variants, giving $280(1+5)=1,680$ training graphs. Validation and test subjects remain unaugmented. These are 1,680 graph instances from 280 people, not 1,680 independent subjects.

The perturbation's symmetry, clipping, and correlation-matrix constraints are not recorded. Accordingly, it is a graph augmentation, not a validated simulation of physiological variability. Ensembling and attention also change this configuration, so the observed improvement does not isolate augmentation's effect.

\subsection{Broader-cohort models}
The broader analysis compares an RBF-kernel SVM, logistic regression, a multilayer perceptron (MLP), and BrainGAT. Vectorized models use harmonized connectivity features; the logistic-regression result specifically uses tangent-space features, and the MLP result uses correlation features. The exact SVM feature variant is not separately identified in the summary.

BrainGAT uses a graph attention branch with two heads and a hidden dimension of 64, paired with a parallel GCN branch. Its graphs retain the top 15\% of absolute pairwise correlations among 111 regions. This is a separate model from the five-member curated ensemble. Edge-count implications are examined in Section~\ref{sec:representation}; the exact directed-edge and self-loop conventions are not recorded.

\subsection{Post-hoc interpretation}
Gradient-based saliency summarizes regional influence, and GNNExplainer~\cite{ying2019gnnexplainer} is used to identify influential graph features and connections. The available regional summary is labeled as GCN feature importance. We therefore distinguish that displayed result from an independent validation of the final GAT ensemble. Explainer masks can motivate hypotheses for further study but require stability and fidelity evaluation before being interpreted as reliable model explanations.
\section{Results and Analysis}
\subsection{Configuration-level performance}
Table~\ref{tab:curated} preserves the reported metrics. The AAL GCN reaches 73.3\% accuracy and AUC 0.74; the MSDL configuration reaches 84.0\% and 0.84; the GAT ensemble reaches 95.0\% and approximately 0.98. The differences in reported accuracy are 10.7 percentage points between the two GCN configurations and 11.0 points between the MSDL GCN and the ensemble. These are descriptive contrasts between complete pipelines.

\begin{table}[t]
\centering\small
\caption{Reported curated-cohort results. All values are retained from the experiment report. The 84.0\% accuracy requires clarification of its aggregation or denominator; precision/recall averaging conventions are not specified.}
\label{tab:curated}
\begin{tabular}{lrrrr}
\toprule
Configuration & Acc. (\%) & Prec. & Rec. & AUC \\
\midrule
AAL GCN & 73.3 & .73 & .73 & .74 \\
MSDL GCN & 84.0 & .84 & .84 & .84 \\
GAT ensemble & 95.0 & .95 & .95 & $\approx$.98 \\
\bottomrule
\end{tabular}
\end{table}

The 84.0\% intermediate accuracy cannot be reconciled with an integer numerator over a single 60-subject test denominator at the displayed precision ($\frac{50}{60} \approx 83.33\%$, $\frac{51}{60} = 85.00\%$). Its aggregation remains unresolved, likely reflecting an unweighted cross-validation split average or probabilistic thresholding across sub-splits. We preserve the reported value for transparent cataloging but do not use the 10.7-point contrast to claim an isolated atlas gain. All reported Area Under the ROC Curve (AUC) scores were computed from continuous predicted softmax probabilities rather than thresholded binary decisions.

\subsection{Broader-cohort comparison}
Table~\ref{tab:broad} presents all four reported broader-cohort results. The SVM has the highest mean accuracy (69.37\%), while the MLP has the highest AUC (0.7575). BrainGAT reaches 58.26\% accuracy and AUC 0.6268, below all three vectorized baselines. The SVM--BrainGAT difference is 11.11 percentage points in accuracy; the MLP--BrainGAT AUC difference is 0.1307. These comparisons do not support a general advantage for the graph model in this setting.

\begin{table}[t]
\centering\small
\caption{Broader-cohort results under the reported five-fold procedure. The $\pm$ quantities are reproduced as reported dispersion; their precise definition is not specified. The harmonization caveat in Section~\ref{sec:settings} applies to all rows.}
\label{tab:broad}
\begin{tabular}{lrr}
\toprule
Model & Accuracy (\%) & AUC \\
\midrule
SVM (RBF) & $69.37\pm3.12$ & .7508 \\
Logistic regression & $68.89\pm1.63$ & .7391 \\
MLP & $68.79\pm3.64$ & .7575 \\
BrainGAT & $58.26\pm3.60$ & .6268 \\
\bottomrule
\end{tabular}
\end{table}

The majority class comprises $530/1035\approx51.21\%$ of subjects. This is a cohort-composition reference, not an evaluated fold-level baseline. Reported dispersion is not interpreted as a confidence interval, and statistical significance cannot be inferred without fold-level predictions.

The two settings do not estimate the same prediction problem with the same model. Their different preprocessing, atlases, architectures, regularization, and splits preclude assigning the score contrast to cohort size or site heterogeneity. The defensible observation is narrower: the recorded broader-cohort comparison does not favor BrainGAT, even though the curated comparison favors a different GAT ensemble. A conventional classifier is therefore a necessary comparator in interpreting this study's graph-model results.

\subsection{Regional attribution}
\begin{figure}[t]
\centering
\includegraphics[width=\linewidth]{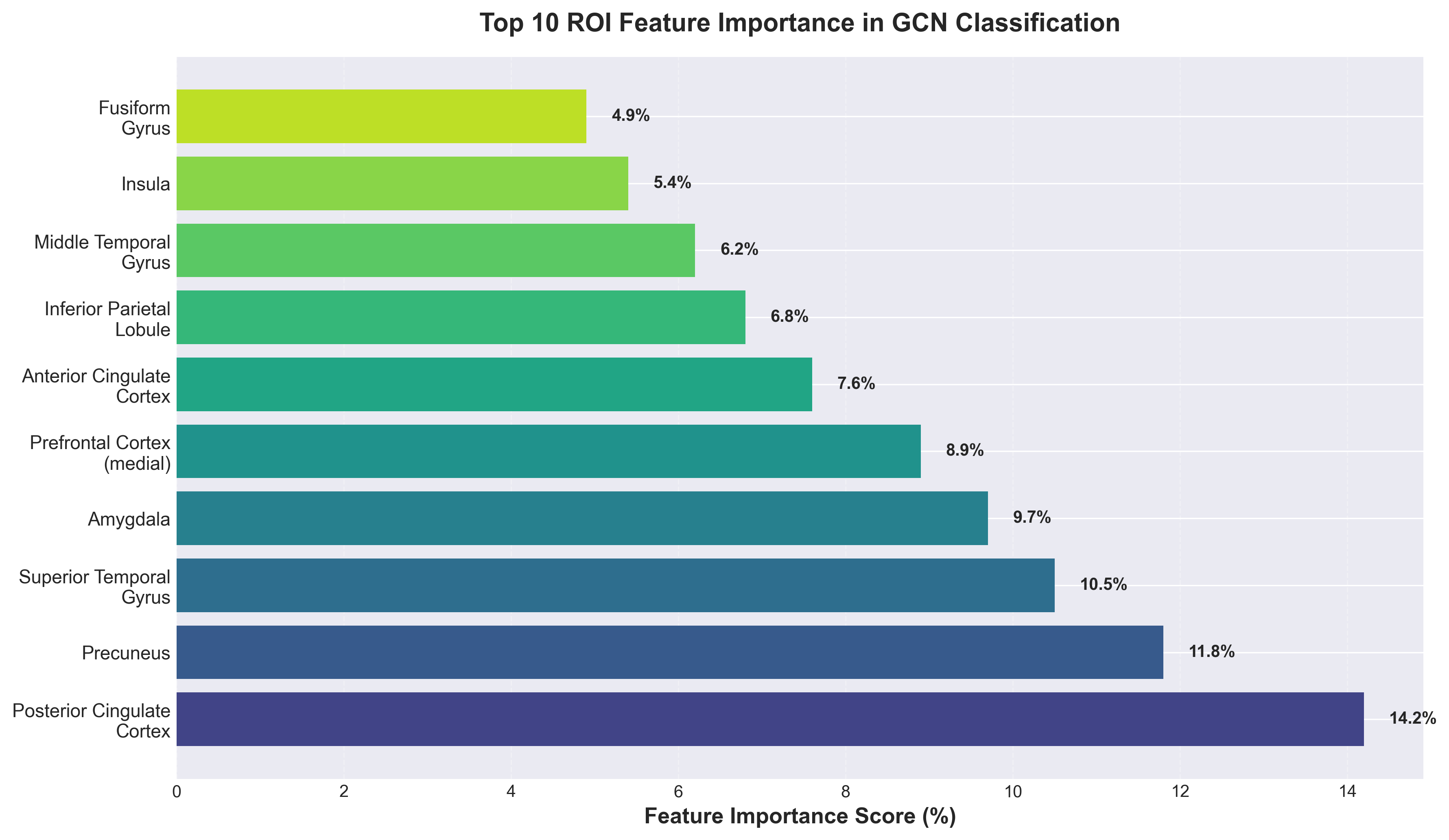}
\caption{Reported GCN regional feature-importance summary. Importantly, these saliency attributions are derived exclusively from the baseline single-model AAL GCN and do not represent the 95.0\% GAT ensemble. The posterior cingulate cortex (14.2\%) and precuneus (11.8\%) suggest classifier reliance on default-mode-network regions, but this reflects internal model attribution rather than validated clinical biomarkers.}
\label{fig:saliency}
\end{figure}

Figure~\ref{fig:saliency} reproduces the regional importance summary. Posterior cingulate cortex and precuneus receive reported scores of 14.2\% and 11.8\%, respectively. These regions provide a plausible connection to default-mode-network hypotheses~\cite{kennedy2008functional,hahamy2015idiosyncratic}. The percentages are attribution scores in the supplied summary, not effect sizes, diagnostic sensitivities, or probabilities that regions are pathological.

No quantitative agreement measure between saliency and GNNExplainer, attribution uncertainty across seeds, or label-randomization control is reported. Consequently, we do not interpret plausible anatomical labels as evidence that site, motion, age, or sex confounding has been excluded. Attribution should be evaluated jointly with predictive performance and with perturbation tests that assess whether highlighted features actually influence the trained classifier.

\subsection{Representation size and confounded atlas comparisons}
\label{sec:representation}
Changing from AAL to MSDL reduces the number of regions from 116 to 39. For a symmetric connectivity matrix without diagonal entries, the number of possible region pairs is
\begin{equation}
 d(R)=\frac{R(R-1)}{2}.
\end{equation}
Table~\ref{tab:dimensions} gives the resulting sizes. The AAL representation has 6,670 possible pairs and MSDL has 741, a ratio of approximately 9.00. This is an arithmetic property of the representations, not a measurement of effective degrees of freedom or the model's parameter count.

\begin{table}[t]
\centering\small
\caption{Derived representation sizes, excluding self-connections. Dense correlation pairs are not the same as retained graph edges. Counts do not establish which representation is more predictive.}
\label{tab:dimensions}
\begin{tabular}{lrr}
\toprule
Parcellation & Regions & Possible pairs \\
\midrule
AAL & 116 & 6,670 \\
MSDL & 39 & 741 \\
Harvard--Oxford & 111 & 6,105 \\
\bottomrule
\end{tabular}
\end{table}

At 20\% density, AAL retains approximately 1,334 undirected connections under a no-diagonal convention. If the same density were applied to MSDL, it would retain approximately 148; this latter value is a conditional calculation, since the MSDL threshold is not explicitly documented. At the documented 15\% density, Harvard--Oxford gives approximately 916 undirected pairs, or 1,832 directed entries before self-loops. This helps interpret the recorded approximate count of 1,850 edges, but does not reconstruct the missing edge convention.

The ninefold change in candidate pairs offers an alternative interpretation to a purely anatomical-versus-functional explanation: graph size and input complexity change at the same time as regional meaning. The MSDL GCN additionally changes hidden-layer depth and normalization. The experiment therefore supports a configuration-level association. It cannot determine whether functional coherence, reduced representation size, network capacity, or their interaction explains the difference.

\subsection{Finite-test-set sensitivity}
\label{sec:finite}
For $n$ test subjects, changing one prediction changes accuracy by $100/n$ percentage points. At $n=60$, this increment is 1.67 points. Thus an apparent 95.0\% result is compatible with 57 correct classifications, and one additional error would reduce accuracy to 93.3\%. This calculation illustrates sensitivity to individual cases; it does not recover the missing predictions.

As a conditional uncertainty calculation, if the 95.0\% result represents 57 successes among 60 independent test subjects for a fixed model, a 95\% Wilson binomial interval is approximately 86.3--98.3\%. For estimated proportion $\hat p$ and $z=1.96$, its endpoints are
\begin{equation}
\frac{\hat p+z^2/(2n)\;\pm\;z\sqrt{\hat p(1-\hat p)/n+z^2/(4n^2)}}{1+z^2/n}.
\end{equation}
This is not a repeated-training confidence interval and does not account for model selection, shared site effects, or the uncertainty in the recorded evaluation procedure. It shows why the headline percentage alone cannot establish reliable clinical performance. The 73.3\% result is similarly compatible with 44 of 60 correct, while 84.0\% requires an aggregation or denominator other than the stated single-test-set interpretation.

\section{Discussion}
\paragraph{Three important results at different levels.}
The experiment demonstrates a positive ensemble configuration but does not specify the precise contribution of each component.

Instead of a validated transfer failure of the curated ensemble, the broader analysis indicates a negative graph-model result within its recorded protocol. The saliency summary explains model attribution instead of the discovery of a clinical biomarker. These distinct levels make negative and positive findings jointly interpretable without treating the pipelines as similar.

\paragraph{Architecture choice and the evaluation objective.}
The SVM ranks first by accuracy and the MLP first by AUC in the wider analysis. Accuracy is measured at an operating threshold. It has been demonstrated that AUC summarizes score ordering across thresholds. The different winners are consistent with various evaluation objectives and do not show contradictory outcomes. Neither metric establishes clinical usefulness or calibration. However, the absence of a graph-model advantage in this setting also argues for retaining simple baselines when graph representations are introduced, instead of assuming that their rich inductive biases improve classification.

\paragraph{Implications for neural classification.}
There are three connected practical decisions: how these regions define the space for input, how the test partitions are used to define the generalization claim, and how explanation is used to define the claim about this model behavior. Furthermore, these problems extend beyond ASD whenever neural data are classified across subjects, acquisition devices, and sessions. The evidence concerns the held-out subjects under the given evaluation settings. No site-disjoint or prospective evaluation was carried out. The interpretation of anatomical data cannot go beyond the generalization boundary.

\section{Limitations and Clinical Scope}
\label{sec:limitations}
\paragraph{Unresolved evaluation provenance.}
The 84.0\% denominator anomaly, exact training-fold isolation of ComBat (which incorporated diagnosis as a covariate), and checkpoint selection protocols remain critical documentation constraints. While we report the original published metrics faithfully, the lack of site-disjoint prospective splits and the potential for information leakage during batch harmonization prevent these scores from being interpreted as verified clinical generalization.

\paragraph{Cohort and uncertainty.}
Both analyses draw on ABIDE-I. Subject overlap is not established, and the reported 17 versus 20 site groups lack a reconciled identifier mapping. Neither constitutes independent external validation of the other. Detailed demographic matching, per-site errors, and repeat-seed variability are unavailable. The conditional binomial calculation in Section~\ref{sec:finite} addresses only a narrow sampling interpretation and cannot substitute for these measurements.

\paragraph{Interpretability and clinical use.}
Saliency normalization and cross-subject aggregation are not fully specified, and no quantitative agreement or stability analysis between explanation methods is available. Plausible default-mode attributions therefore remain hypotheses. The study predicts dataset labels and does not establish diagnostic utility, calibration, or subgroup fairness. A balanced research cohort also does not determine predictive value in a clinical referral population. Neural data and linked demographics require handling under the source dataset's access and redistribution conditions. Clinical decisions should not be based on this exploratory classifier.

\section{Conclusion}
The two ABIDE-I analyses show why graph-based ASD classification must be interpreted at the level of the full representation and evaluation pipeline. A curated ensemble reports high accuracy, whereas a separate broader-cohort graph model ranks below conventional baselines. Analysis of the existing summaries quantifies the ninefold difference in candidate connectivity pairs between AAL and MSDL and the 1.67-point accuracy increment on a 60-person test set. These calculations sharpen the interpretation of the reported results without adding experiments or assigning a causal explanation to confounded comparisons. The unresolved metric aggregation and harmonization procedures limit clinical conclusions. The resulting evidence supports careful configuration comparison and hypothesis generation, rather than a claim of validated ASD diagnosis.
\section*{LLM Usage Statement}
ChatGPT (OpenAI) assisted with manuscript drafting and editing,
consistency checks and LaTeX debugging.
Responsibility for the scientific claims, data, analyses, and final
submission remains with the authors.
\FloatBarrier
{\small\bibliographystyle{ieeenat_fullname}\bibliography{references}}
\end{document}